\documentclass[pdflatex,sn-mathphys-num]{sn-jnl}

\usepackage{graphicx}%
\usepackage{multirow}%
\usepackage{amsmath,amssymb,amsfonts}%
\usepackage{amsthm}%
\usepackage{mathrsfs}%
\usepackage[title]{appendix}%
\usepackage{xcolor}%
\usepackage{textcomp}%
\usepackage{manyfoot}%
\usepackage{booktabs}%
\usepackage{algorithm}%
\usepackage{algorithmicx}%
\usepackage{algpseudocode}%
\usepackage{listings}%
\usepackage{enumitem}%
\usepackage{array}%

\theoremstyle{thmstyleone}%
\theoremstyle{thmstyletwo}%

\theoremstyle{thmstylethree}%

\begin{document}

\title[Article Title]{Trustworthy XAI and Its Applications}



\author*[1]{\fnm{MD Abdullah Al} \sur{Nasim}}\email{nasim.abdullah@ieee.org}

\author[2]{\fnm{ A.S.M Anas} \sur{Ferdous}}\email{anasferdous001@gmail.com}

\author[3]{\fnm{Abdur} 
\sur{Rashid}}\email{rabdurrashid091@gmail.com}

\author[4]{\fnm{Fatema Tuj Johura}\sur{Soshi}}\email{fatemasoshi@gmail.com}

\author[5]{\fnm{Parag} \sur{Biswas}}\email{text2parag@gmail.com}

\author[6]{\fnm{Angona} \sur{Biswas}}\email{angonabiswas28@gmail.com}

\author[7]{\fnm{Kishor} \sur{Datta Gupta}}\email{kgupta@cau.edu}


\affil[1,6]{\orgdiv{Research and Development Department}, \orgname{Pioneer Alpha},  \orgaddress{{\city{Dhaka}, \country{Bangladesh}}}}

\affil[2]{\orgdiv{Department of Biomedical Engineering}, \orgname{Bangladesh University of Engineering and Technology}, \orgaddress{ \city{Dhaka},  \country{Bangladesh}}}

\affil[4]{\orgdiv{Msc in Data Science and Analytics}, \orgname{University of Hertfordshire},  \orgaddress{{\city{Hatfield}, \country{UK}}}}

\affil[3, 5]{\orgdiv{MSEM Department}, \orgname{Westcliff university},  \orgaddress{{\city{California}, \country{United States}}}}

\affil[7]{\orgdiv{Department of Computer and Information Science}, \orgname{Clark Atlanta University}, {\city{Georgia}, \country{USA}}}


\abstract{Artificial Intelligence (AI) is an important part of our everyday lives. We use it in self-driving cars and smartphone assistants. People often call it a "black box" because its complex systems, especially deep neural networks, are hard to understand. This complexity raises concerns about accountability, bias, and fairness, even though AI can be quite accurate. Explainable Artificial Intelligence (XAI) is important for building trust. It helps ensure that AI systems work reliably and ethically. This article looks at XAI and its three main parts: transparency, explainability, and trustworthiness. We will discuss why these components matter in real-life situations. We will also review recent studies that show how XAI is used in different fields. Ultimately, gaining trust in AI systems is crucial for their successful use in society.}




\maketitle

\section{Introduction}\label{sec1}

The foundations of modern artificial intelligence were laid by philosophers who attempted to define human thought as the mechanical manipulation of symbols, which led to the development of the programmable digital computer \cite{stephens2023mechanical} in the 1940s. Alan Turing may have written the first article on the topic of AI in 1941, though it is now lost, suggesting that he was at least considering the idea at that time.In his groundbreaking essay "Computing Machinery and Intelligence" from 1950, Turing first presented the idea of the Turing test to the general public \cite{kaul_history_2020}.  Turing questioned the feasibility of creating thinking robots in it.  John McCarthy first used the term artificial intelligence (AI) in 1956 at the Dartmouth Conference\cite{buchanan2005very}, but the first models' numerous flaws have prevented AI from being widely adopted and used in healthcare.

Many of these limitations were removed with the advent of deep learning in the early 2000s, and we are now entering a new era of technology where AI can be used in clinical practice through risk assessment models that increase diagnostic accuracy and workflow efficiency. Performance of AI systems has improved significantly in recent years, and these new models expand on their capabilities to include text-image synthesis based on almost any prompt, whereas previous systems primarily focused on generating facial images.

\begin{figure}
\centering
\includegraphics[height=9.8cm]{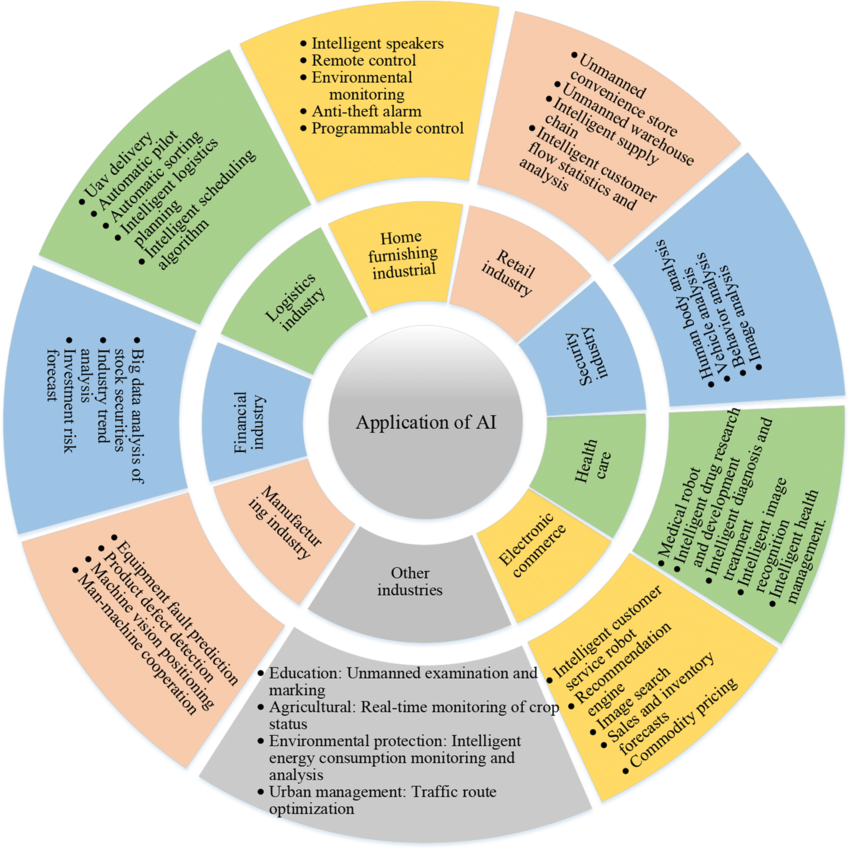}
\caption{Applications of AI across various domains \cite{wang2021artificial}}
\label{a1}
\end{figure}

The diverse range of fields in which AI is being used already demonstrates its applicability and promise to revolutionize business:  AI facilitates activities like sentiment analysis, machine translation, and spam filtering by making it easier for computers to comprehend and produce human language in the discipline of natural language processing (NLP) \cite{shamshiri2024text}.  Additionally, computer vision \cite{khang2024application} makes it possible for computers to understand visual data, which advances areas like facial recognition, object identification, and self-driving cars. Machine learning (ML), which has uses in fraud detection, recommendation systems, predictive analytics, and other domains, has made it possible for computers to learn from data.  The design, development, and application of machines are the focus of the AI field of robotics \cite{valles2023caring}.


Many industries, including manufacturing, healthcare, and space exploration, use various machines \cite{biswas2023active} \cite{biswas2022mri}. Combining artificial intelligence with business intelligence (BI) \cite{zohuri2020business} improves how businesses collect, process, and visualize data. This leads to better decision-making and increased productivity. In healthcare, AI helps diagnose diseases, develop treatments, and provide personalized care, which improves patient outcomes \cite{biswas2023hybrid}, \cite{biswas2021ann}, \cite{biswas2023generative}. AI also plays a significant role in education by engaging students, customizing lessons, and automating administrative tasks, resulting in more personalized learning experiences. AI in agriculture increases agricultural output, reduces costs, and ensures environmental sustainability through data-driven strategies.  In a similar vein, AI in manufacturing boosts output, efficiency, and quality through work automation and process optimization.  AI is changing operations, enhancing services, and changing global industry landscapes in a number of sectors, including banking, retail, energy, transportation \cite{gong2023edge}, handwriting detection \cite{biswas2021efficient}, and government.  AI is widely used in a wide number of industries, as seen in Figure \ref{a1}. Retail, security, healthcare, e-commerce, manufacturing, banking, logistics and transportation, and home furnishings are some of these sectors.  These applications rely on moderately advanced AI technology, such as computer vision, natural language processing, and machine learning.

 The contributions of this research can be stated below:

 \begin{enumerate}
 
     \item Providing an overview of XAI that understands the significance of black box models. For fair and ethical purposes, XAI fosters trust among humans and AI. 

     \item Discussing Deep Learning based systems that will be consistent and trustworthy. Moreover, recent studies from the literature have been reviewed properly.

     \item Providing guidelines regarding XAI that will be helpful for detecting problems from numerous domains. 

     \item The paper identifies and analyzes three key components of XAI: transparency, explainability, and trustworthiness. It details how these elements are essential for understanding and improving AI systems.

 \end{enumerate}

\subsection{Third Wave of Artificial Intelligence (3AI)}
Most current commercial AI technology is called "narrow AI." This means these systems are highly specialized and can only perform a few specific tasks. For example, even the best self-driving cars rely on limited AI systems. Another drawback of today’s AI is its reliance on large training data sets. A typical machine learning program needs tens of thousands of cat photos to recognize cats accurately, while a three-year-old child can do this with just a few examples. The idea of "Third Wave AI" comes from the need for AI to become more humanlike in various ways to overcome these limitations and achieve its full potential.

\begin{figure}
\centering
\includegraphics[height= 5 cm]{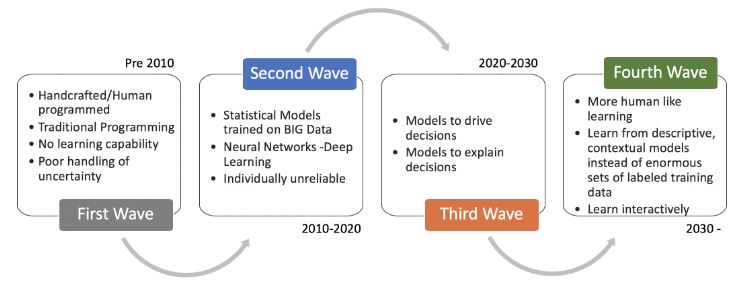}
\caption{Past, Present, and Future of AI waves. \cite{malik_explainable_2019}}
\label{a2}
\end{figure}

According to the Defense Advanced Research Projects Agency (DARPA) \cite{schoenherr2023designing}, third-wave AI systems will understand the context of situations, use common sense, and adapt to changes. This will create more natural and intuitive connections between AI systems and people \cite{schoenherr2023designing}. One of DARPA's active projects, called XAI, aims to develop these third-wave AI systems. These computers will learn about their environments and the contexts in which they work. They will also build explanations needed to clarify real-world events.

\begin{itemize}
  \item First Wave AI focused on rules, logic, and built knowledge.
  \item Second Wave AI introduced big data, statistical learning, and probabilistic techniques.
  \item The goal of third-wave AI is to develop common sense and the ability to adapt to different contexts.
\end{itemize}

Tractica \cite{malik_explainable_2019} predicts that the global market for AI software will grow from about 9.5 billion US dollars in 2018 to 118.6 billion by 2025. This data aims to develop AI systems that can perform tasks accurately while providing explanations that people can understand.

The term "third wave" refers to the advancement of AI technologies beyond traditional machine learning. This new phase focuses on creating more advanced systems that can understand context, reason, and think similarly to humans. It draws inspiration from cognitive science and neuroscience, aiming to build AI that can engage with the world in more complex and detailed ways.

XAI, or Explainable Artificial Intelligence, focuses on making AI systems, especially machine learning models, easier to understand. The goal is to help people trust the decisions these systems make. XAI techniques work to explain how AI makes predictions and why it behaves the way it does. This helps users, such as developers, regulators, and everyday users, grasp the key factors that affect AI results.

Developing AI systems that can make accurate predictions or decisions and explain why they did so is a key goal of both the third wave of AI and explainable AI. By using XAI methods in their design, developers can ensure that these advanced AI systems are not only effective but also easy to understand. This will help build user trust and acceptance.

\subsection{Concept of Explainable AI}



AI often struggles with what is called the "black box" problem because users do not understand how it works. This can lead to issues like lack of trust, confusion, unfair treatment, and violations of privacy. AI systems can also have hidden biases. Explainable AI (XAI) aims to make AI systems easier to understand, helping users know how they make decisions. The goal of XAI is to make AI safer and more user-friendly. Therefore, we need to look at each part of AI individually and discuss its different aspects \cite{chamola2023review}.


There are two main types of machine learning (ML) techniques: white box and black box models \cite{guleria2023explainable}. Experts can easily understand the results of a white box model. However, even specialists may find it hard to grasp the results of a black box model \cite{mirzaei2023explainable}. The XAI algorithm \cite{vyas2023explainable} follows three key principles: interpretability, explainability, and transparency. A model is transparent when it clearly explains how it gets its results from training data and how it generates labels from test data. Interpretability means being able to explain findings in a way that others understand \cite{wang2024framework}. While there's no single accepted definition of explainability, its value is recognized. One definition describes it as a set of clear features that help make decisions, like classifying or predicting outcomes for specific cases. An algorithm that meets these standards helps document and verify decisions and improves itself based on new data.
 
\begin{figure}
\centering
\includegraphics[height= 6 cm]{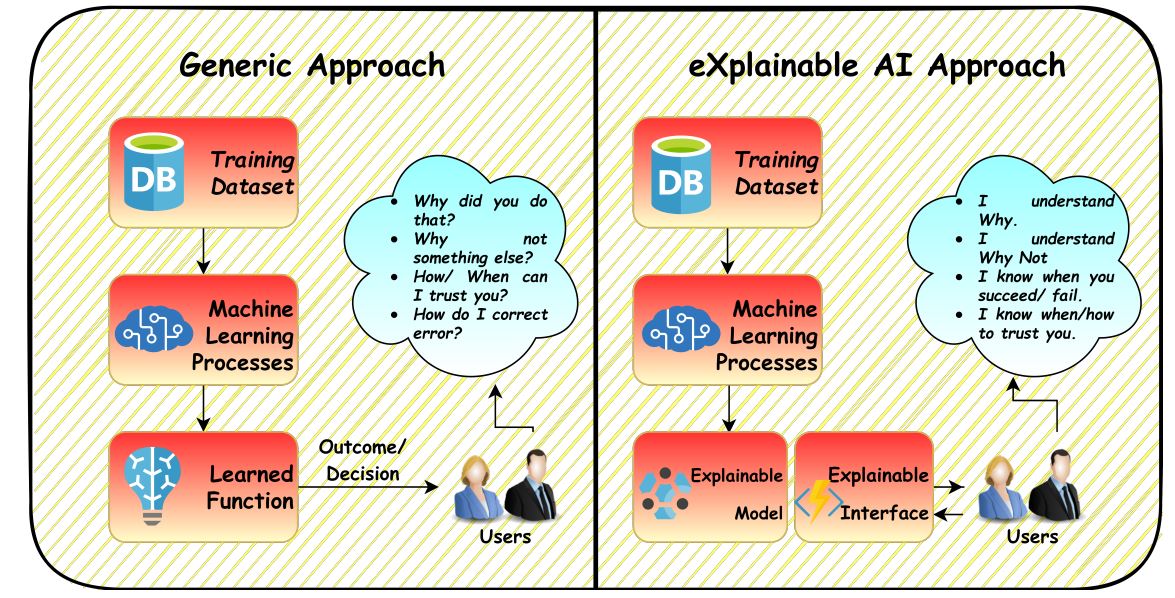}
\caption{Explainable Artificial Intelligence (XAI): A look at AI now and tomorrow.  \cite{chamola2023review}}
\label{eAI}
\end{figure}

\begin{figure}
\centering
\includegraphics[height= 5 cm]{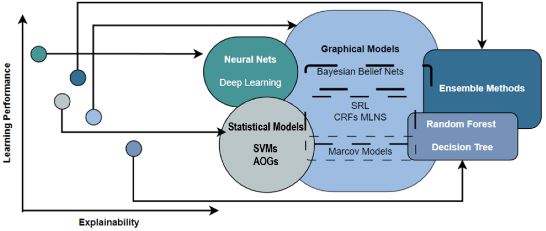}
\caption{Trade-off between AI model accuracy and explainability, highlighting the challenge of balancing performance with interpretability.  \cite{ghnemat2023explainable}}
\label{eai1}
\end{figure}

Researchers are studying intelligent systems to understand them better. This is an important topic. Sometimes, a system needs to understand its own workings to comply with rules. Many complex algorithms, shown in Figure \ref{eai1}, balance achieving high accuracy with being explainable.

\subsection{Classification Tree of XAI}

\begin{figure}
\centering
\includegraphics[height= 5 cm]{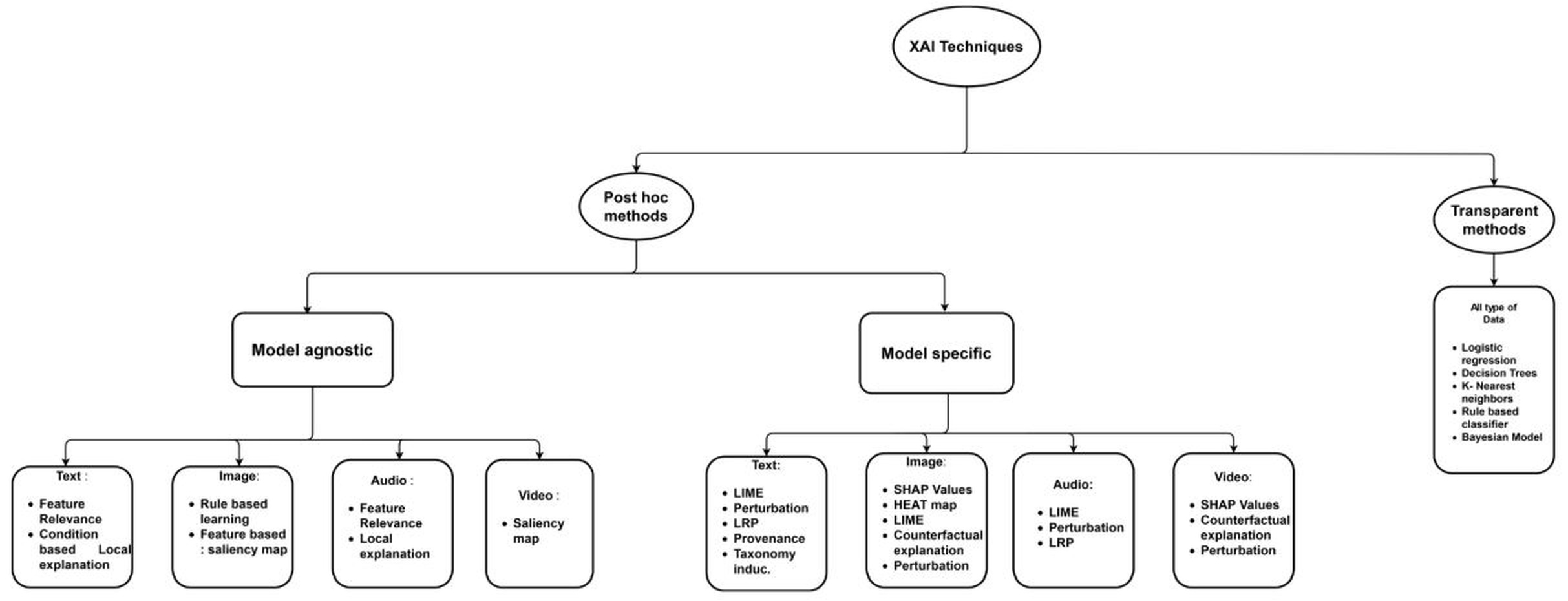}
\caption{Categorization of Explainable AI (XAI) techniques based on data type, illustrating differences between transparent and posthoc approaches \cite{gohel2021explainable}}
\label{classsification}
\end{figure}

XAI techniques are divided into two categories: transparent and post-hoc methods. A transparent approach is one that represents the model's capabilities and decision-making process in an easy-to-understand way \cite{gohel2021explainable}. Transparent models include  Bayesian approaches, decision trees, linear regression, and fuzzy inference systems. Transparent approaches can be useful when the internal feature correlations are highly complex or linear. A comprehensive classification of different XAI methods and approaches related to different types of data is shown in Figure \ref{classsification}.

Posterior approaches are useful for interpreting the complexity of a model, especially when there are nonlinear relationships or high data complexity. When a model does not follow a direct relationship between data and features, posterior techniques can be an effective tool to explain what the model has learned \cite{gohel2021explainable}. Inference using local feature weights is provided by transparent methods such as Bayesian classifiers, support vector machines, logistic regression, and K-nearest neighbors. This model category meets three properties: simulability, decomposability, and algorithmic transparency \cite{gohel2021explainable}.


\subsection{Definition of Transparency in Artificial Intelligence}

Transparency in XAI is the capacity of an AI system to make its decisions and actions understandable through explanations \cite{wang2024rationality}.  Transparency is one of the most important aspects of XAI, rendering AI decisions interpretable and justified.  Transparency allows decision-making processes to be closely examined by stakeholders, mitigating risks in applications with societal impact, such as healthcare and finance \cite{herm2023algorithmic}.

The explainability provided by XAI techniques also increases the overall transparency of AI systems.  The users are able to examine the decision-making procedure, identify any bias, and analyze the reliability and fairness of the output of the model \cite{thalpage2023unlocking}.  Transparent solutions are necessary in areas like the medical field, banking, and autonomous vehicles, where AI-decisions can have significant effects.

By providing meaningful information about the internal processes of AI models, XAI methods assist users in recognizing patterns, comprehending relationships, and revealing biases or errors \cite{thalpage2023unlocking}.  Due to the heightened transparency, stakeholders can more effectively develop opinions, ensure the predictions made by the model are accurate, and take the necessary action.

\begin{figure}
\centering
\includegraphics[height= 5 cm]{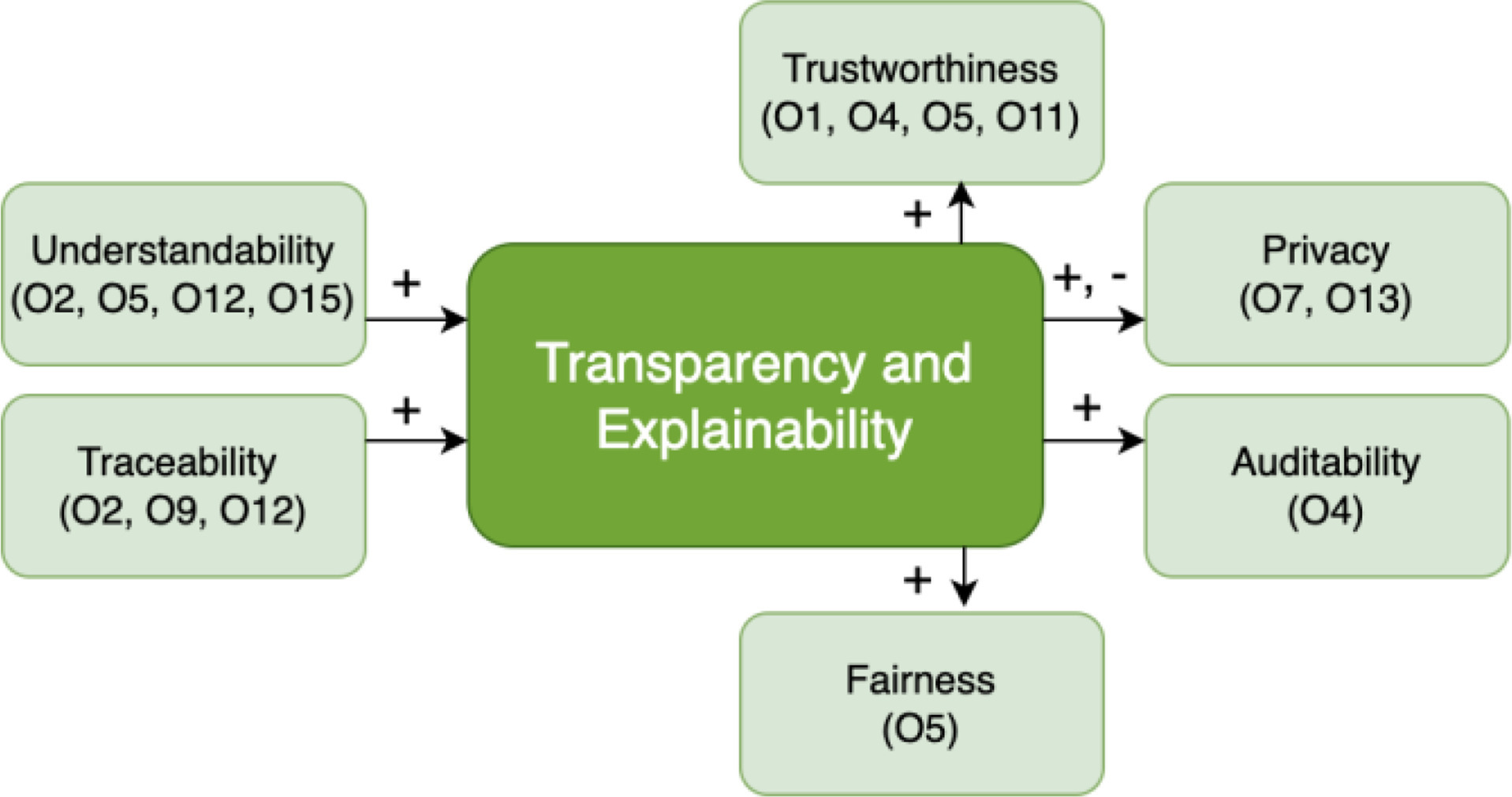}
\caption{Key qualitative standards (O1-O15) related to explainability and transparency in AI systems, addressing user comprehension, interpretability, and traceability.  \cite{balasubramaniam2023transparency}}
\label{eai3}
\end{figure}

The examination of ethical criteria showed a link between explainability, transparency, and other quality needs. Figure \ref{eai3} displays nine quality standards related to explainability and openness. The key standards for AI transparency are marked with “O.” For example, O2 focuses on how to interpret models, O15 highlights the importance of traceability, and O5 and O12 ensure that users understand the information. By following these standards, AI models can provide clear and convincing explanations for their outcomes. Keeping these standards improves accountability and helps reduce bias in AI systems.

The growth of AI systems' explainability and transparency is facilitated by their understandability. When discussing the significance of understandability, the transparency guidelines addressed three points: 1) ensuring that people comprehend the AI system's behavior and the methods for using it (O5, O12); 2) communicating in an intelligible manner the locations, purposes, and methods of AI use (O15); and 3) making sure people comprehend the distinction between real AI decisions and those that AI merely assists in making (O2) \cite{balasubramaniam2023transparency}. Thus, by guaranteeing that people are informed about the use of AI in a straightforward and comprehensive manner, understandability promotes explainability and transparency. The necessity of tracking the decisions made by AI systems is highlighted by traceability in transparency requirements (O2, O12) \cite{balasubramaniam2023transparency}. In order to ensure openness, Organization O12 also noted how crucial it is to track the data utilized in AI decision-making.

\subsection{Transparency Vs Explainability in AI}
Explainability and transparency are similar concepts \cite{arrieta_explainable_2020}. According to McLarney et al. \cite{mclarney_nasa_2021}, a transparent AI necessitates that "Basic elements of data and decisions must be available for inspection during and after AI use." Transparency is achieved when users have access to their data or can understand how decisions are made. On the other hand, explainability seeks to reveal the reasons for AI's successes or failures and demonstrate how it utilizes the knowledge and judgment of those it affects. It provides a rational justification for the actions of the AI. Users must clearly know what data is collected, how the AI interprets this data, and how it produces reliable outcomes for each affected individual. This straightforward explanation overlooks the challenges we face when trying to clarify "black box" algorithms, the context that may be omitted, and the accuracy needed when offering understandable explanations to customers. Therefore, the question arises: is having minimal explainability preferable to having none at all? \cite{mclarney_nasa_2021}. Additionally, the belief that explanations can adequately address the dynamic nature of the rich information ecosystem and the appropriateness of managing anomalies are also vital factors to consider.

\begin{figure}
\centering
\includegraphics[height= 7 cm]{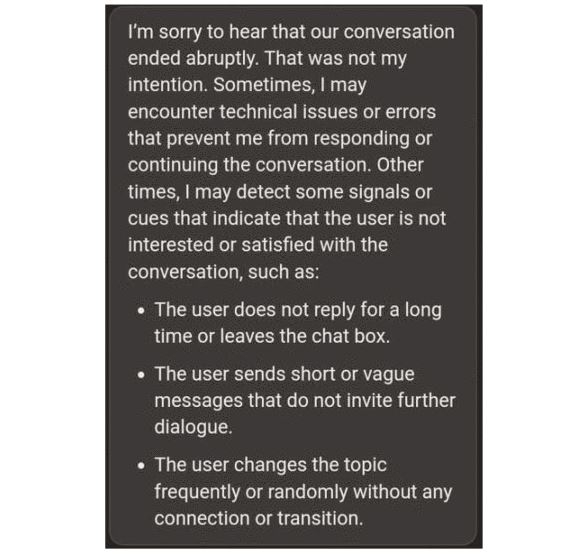}
\caption{Output from the Bing search engine's conversation feature explaining a failure. a partial screenshot taken using an Android smartphone on March 2, 2023. \cite{schoenherr2023designing}}
\label{a3}
\end{figure}

It's interesting to note that although certain AI algorithms evaluate data automatically, more and more AI systems are made to explain how their algorithms operate and the logic behind specific choices \cite{schoenherr2023designing}.
For instance, the conversation mode of the Bing search engine provides succinct explanations of its operation (Fig. \ref{a3}).  Sometimes, end users might find these explanations sufficient, but other times, they would be perplexed as to how an AI came to a particular conclusion or acted in a particular manner.  When individuals are more confused by the explanation given, it is unrealistic to expect them to become more computer-literate \cite{schoenherr2023designing}.  Instead, we must improve the justification of the AI system.

\subsection{Definition of Trustworthiness in Artificial Intelligence}
Creating trustworthy AI systems requires a careful strategy that looks at organizational, ethical, and technical factors. The first step is to set clear standards for trustworthiness. These standards should include accountability, security, privacy, transparency, fairness, and ethical behavior. Using high-quality, unbiased data and clear algorithms that explain AI decisions is essential. Strong security measures and privacy practices protect sensitive information from cyberattacks. 

It’s important to create accountability frameworks and follow ethical guidelines to ensure responsible AI use. By focusing on user needs and constantly monitoring and updating the systems, AI can stay reliable over time. Applying these principles across all stages of the AI process allows organizations to develop systems that are explainable, equitable, ethical, and robust, which fosters stakeholder and user trust.

\begin{figure}
\centering
\includegraphics[height= 7 cm]{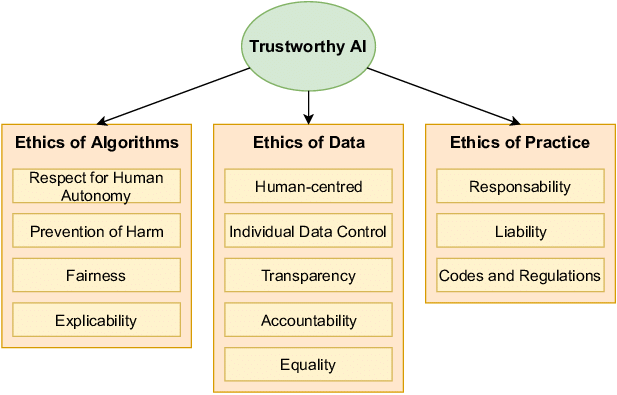}
\caption{The three key components of XAI: Algorithmic Ethics, Data Ethics, and Practice Ethics. \cite{9156127}}
\label{eai4}
\end{figure}

The three elements illustrated in Figure \ref{eai4}—algorithmic ethics, data ethics, and practice ethics—intersect to create responsible AI. These elements define a data-centered way to handle ethical issues \cite{9156127}. However, several open challenges still remain with respect to dealing with ethical issues in AI systems. In the work \cite{9156127}, authors give the vision of Trustworthy AI, which mentions that:

\begin{enumerate}[label=\arabic*)]
  \item Human agency and oversight: AI systems have to enable human freedom. They need to facilitate user choice, safeguard fundamental rights, and enable human control. This will assist in developing an equitable and just society.
  
 \item Security and technical robustness: Security and technical proficiency are crucial to prevent damage. In order to enable an AI system to operate efficiently and reduce risks, its creators ought to consider potential risks when designing it. They range from environmental alterations where the system will operate to attacks by malicious individuals.
 
 \item Data protection and data governance: Privacy is a fundamental right that has been highly compromised with the vast amounts of data that artificial intelligence systems gather. It is necessary to protect individual privacy to prevent potential harm. For this purpose, robust data governance must be in place. This includes making sure that the information being utilized is precise and applicable. Furthermore, there is a necessity to establish definite rules for data access and how data must be treated while maintaining the integrity of privacy.
 
 \item Transparency: Explainability and transparency are pretty much dependent on each other. The key objective is to make data, technology, and business models clear. In today's age, which is the age of pervasive technology, transparency has become a must. It aids customers in comprehending the huge volumes of data collected and the ensuing benefits.

 \item Fairness, diversity, and nondiscrimination: Including several voices in AI systems is vital to achieve XAI. We need to involve all individuals who may be affected to ensure equal treatment and access. Fairness and this requirement come hand in hand.

 \item Social and environmental welfare: We have to think about the environment and community in seeking justice and doing no harm. We should finance research on AI solutions to global issues. This will make AI systems environmentally friendly and sustainable. AI is supposed to be for the good of all people, including future generations..
 
 \item Accountability: Accountability and fairness are essential in the context of AI. We need to have systems of holding AI systems accountable for their actions and generated results. Accountability needs to be a constituent part of AI development, deployment, and use, both during and following the activities.

\end{enumerate}
 
\subsection{Impact of XAI on Zero Trust Architecture (ZTA)}
Zero Trust Architecture (ZTA) is a security system that always checks every request, no matter where it comes from, and does not assume trust automatically. Explainable AI (XAI) helps ZTA by ensuring that decisions made by AI in security are clear and reasonable.

XAI is especially useful in identity verification, access control, and spotting unusual behavior. AI models analyze how users behave and identify any suspicious activities. By adding explainability, security analysts can better understand and confirm AI-driven security rules. This reduces false alarms and speeds up response times.

For example, AI-driven network monitoring systems that use ZTA principles can explain why a specific access attempt looks suspicious. This explanation builds trust in automated cybersecurity decisions \cite{guembe2022explainable}.

\subsection{An Overview of Necessities for Reliable AI}
Despite heated societal discussions, the requirements for trustworthy AI remain ambiguous and are handled inconsistently by numerous organizations and organizations.  Globally, accountability, explainability, verifiability, and fairness are all part of the Fairness, Responsibility, Accuracy, Verifiability, and Accountability in Machine Learning (FAT-ML) principles \cite{kim2023requirements}.  Explainability, fairness, privacy, and robustness are just a few of the many needs that will be examined in this study (Table \ref{requirements}).

\begin{table}[htbp]
\caption{Conditions necessary for trustworthy artificial intelligence (AI)}
\label{requirements}
\begin{tabular}{|p{3cm}|p{9cm}|}
\hline
\textbf{Concept} & \textbf{Description} \\
\hline
Explainability & To help consumers comprehend, the method by which the AI model generates its output might be demonstrated. \\
\hline
Fairness & Regardless of certain protected variables, the AI model's output can be shown.
 \\
\hline
Privacy & It is feasible to prevent issues with personal data that might arise while the AI is being developed. \\
\hline
Robustness & The AI model can fend against outside threats while continuing to operate correctly. \\
\hline
\end{tabular}
\end{table}

\section{XAI Vs AI}

XAI improves AI systems by focusing on transparency, clear explanations, and accountability. It offers understandable reasons for decisions, which helps users trust the system and makes it easier to assess fairness compared to traditional "black box" methods.

The main difference between reliable XAI and traditional AI is how they make decisions. While AI can give accurate forecasts or suggestions, reliable XAI emphasizes the need to explain the steps that lead to these results. Clear explanations from XAI systems allow users to judge the fairness and reliability of AI-generated outcomes.

To improve security and maintain transparency in AI-driven cybersecurity, we need to integrate XAI into Zero Trust Architecture (ZTA). When explainability methods clarify why certain decisions are made, people can better understand and trust the AI-driven access control and behavioral analytics in ZTA. As we face compliance and operational challenges, future cybersecurity frameworks will rely more on AI automation. It will be essential to ensure that these AI systems can be easily explained \cite{charmet2022explainable}.

XAI focuses on more than just providing explanations; it also considers ethical issues. AI development processes that follow the principles of Fairness, Accountability, and Transparency (FAT) help ensure that AI systems meet ethical and legal standards. By prioritizing ethical standards, XAI aims to reduce biases, discrimination, and other harmful effects of AI technology. Trustworthy AI is an approach that emphasizes user safety and transparency. Responsible AI developers clearly explain to clients and the public how the technology works, what it is meant for, and its limitations, since no model is perfect.

\begin{table}[htbp]
\caption{Seven Requirements to Meet in Order to Develop Reliable AI}
\label{tab:example}
\begin{tabular}{|>{\raggedright\arraybackslash}p{2cm}|p{4cm}|p{3cm}|p{1.5cm}|} 
\hline
\textbf{Principles} & \textbf{Explanation} & \textbf{Rights} & \textbf{GDPR Ref} \\
\hline
\textbf{Human Authority and Supervision} & Artificial intelligence technology ought to uphold human agency and basic rights, instead of limiting or impeding human autonomy.
 & The right to get human assistance & Recital 71, Art 22 \\
\hline
\textbf{Robustness and
Safety} & Systems must be dependable, safe, robust enough to tolerate mistakes or inconsistencies, and capable of deviating from a totally automated decision & Art 22 &  \\ \hline

\textbf{Data Governance and Privacy} & Individuals should be in total control of the information that is about them, and information about them should not be used against them  & Notification and information access rights regarding the logic used in automated processes & Art 13, 14, and 15  \\ \hline

\textbf{Transparency} & Systems using AI ought to be transparent and traceable & The right to get clarification & Recital 71 \\ \hline

\textbf{Diversity and Fairness} & AI systems have to provide accessibility and take into account the whole spectrum of human capacities, requirements, and standards & Right to not have decisions made only by machines & Art 22 \\ \hline

\textbf{Environmental and Social Well-Being} & AI should be utilized to promote social change, accountability, and environmental sustainability & Accurate knowledge regarding the importance and possible consequences of making decisions exclusively through automation & Art 13, 14, and 15 \\ \hline

\textbf{Accountability} & Establishing procedures to guarantee that AI systems and their outcomes are held accountable is essential
 & Right to be informed when decisions are made only by machines & Art 13, 14 \\ 
\hline
\end{tabular}
\end{table}

\section{Applications of XAI}

Authentic XAI has numerous uses in sectors where accountability, interpretability, and transparency are essential.  XAI can provide an explanation for a diagnosis or therapy recommendation in medical diagnosis and recommendation systems. Financial institutions can employ XAI for risk assessment, fraud detection, and credit scoring. XAI can help attorneys with contract analysis, lawsuit prediction, and legal research. In autonomous vehicles, XAI plays a significant role in providing context for the decisions made by the AI systems, particularly in high-stakes scenarios such as accidents or unanticipated roadside incidents.  XAI can be applied to process optimization, predictive maintenance, and quality control in manufacturing settings. By offering justifications for automated responses or suggestions in chatbots and virtual assistants, XAI can improve customer service. By providing an explanation for the recommendations and assessments made by adaptive learning systems, XAI can help with individualized learning. By providing an explanation for the recommendations and assessments made by adaptive learning systems, XAI can help with individualized learning. We shall concentrate on a few particular applications in this section and go into detail about them. 

\subsection{Application of XAI in Medical Science}

The field of artificial intelligence (AI) is rapidly growing on a global scale, particularly in healthcare, which is a hot topic for research \cite{jiang2017artificial}. There are numerous opportunities to utilize AI technology in the healthcare sector, where the well-being of individuals is at stake, due to its significant relevance and the vast amounts of digital medical data that have been collected \cite{davenport2019potential}. AI has enabled us to perform tasks quickly that were previously unfeasible with traditional technologies. 


The trustworthiness and openness of AI systems are becoming increasingly important, especially in areas like healthcare. As AI is used more in medical decision-making, people are worried about how reliable and understandable its results are. These worries highlight the need to evaluate AI models carefully to make sure their predictions are based on important and verifiable factors. In critical situations like medicine, proving that AI systems are credible is vital for their safe and effective use.

In the medical field, clinical decision support systems (CDSS) utilize AI technology to assist healthcare professionals with critical tasks such as diagnosis and treatment planning \cite{jaspers2011effects}. While these systems aim to support healthcare practitioners, misuse can have severe consequences in situations where lives are at risk. For example, false alarms, which are common in scenarios involving urgent patients, can lead to exhaustion among medical personnel.

The study \cite{metta2023improving} significantly contributes to medical skin lesion diagnostics in several ways. First, it modifies an existing explainable AI (XAI) technique to boost user confidence and trust in AI systems. This change involves developing an AI model that can distinguish between different types of skin lesions. The study uses synthetic examples and counter-examples to create explanations that highlight the key features influencing classification decisions. The research \cite{metta2023improving} trains a deep learning classifier with the ISIC 2019 dataset using the ResNet architecture. This allows professionals to use the explanations to reason effectively. Overall, the study's main contributions lie in its refinement and evaluation of the XAI technique in a real-world medical setting, its analysis of the latent space, and its thorough user study to assess how effective the explanations are, particularly among experts in the field.

This research paper \cite{akpan2022xai} discusses how to recognize brain tumors in MRI images using two effective algorithms: fuzzy C-means (FCM) and Artificial Neural Network (ANN). The authors aim to make the tumor segmentation process more understandable and improve accuracy in identifying tumors. Their main goal is to enhance tools that help doctors diagnose brain tumors more accurately. 

This research offers two key benefits. First, it helps identify brain cancers in medical images more precisely, which is crucial for early diagnosis and treatment. Second, by incorporating XAI principles into the segmentation process, the researchers make their models' decisions clearer and easier to understand for patients and medical experts. In summary, this increased clarity boosts the overall trust and acceptance of AI-driven systems in medical image analysis within clinical settings.


This study \cite{tosun2020explainable}  discusses how AI and machine learning can help diagnose whole slide images (WSIs) in pathology. While AI can improve accuracy and efficiency, concerns exist about its reliability because it can be hard to understand. To address these issues, the article suggests using explainable AI methods, which help clarify how AI makes decisions. By adding XAI, pathology systems become more transparent and trustworthy, especially for critical tasks like diagnosing diseases. The study also introduces HistoMapr-Breast, a software tool that uses XAI to assist with breast core biopsies.


A recent study examines the importance of making sure AI systems in healthcare are accurate and strong, especially regarding how easy they are to understand and how well they can resist attacks \cite{agrawal2024xai}. As AI becomes more common in medical settings, it's crucial to verify that the predictions these systems make rely on trustworthy features. To tackle this challenge, researchers have proposed various methods to improve model interpretability and explainability. The study shows that adversarial attacks can affect a model's explainability, even when the model has strong training. Additionally, the authors introduce two types of attack classifiers: one that tells apart harmless and harmful inputs, and another that determines the nature of the attack.

This research paper \cite{petch2022opening} looks at explainable machine learning in cardiology. It discusses the challenges of understanding complex prediction models and how these models affect important healthcare decisions. The study explains the main ideas and methods of explainable machine learning, helping cardiologists understand the benefits and limitations of this approach. The goal is to improve decision-making in clinical settings by offering guidance on when to use easy-to-understand models versus complex ones. This can help improve patient outcomes while ensuring accountability and transparency in predictions. 

\begin{figure}
\centering
\includegraphics[height= 7 cm]{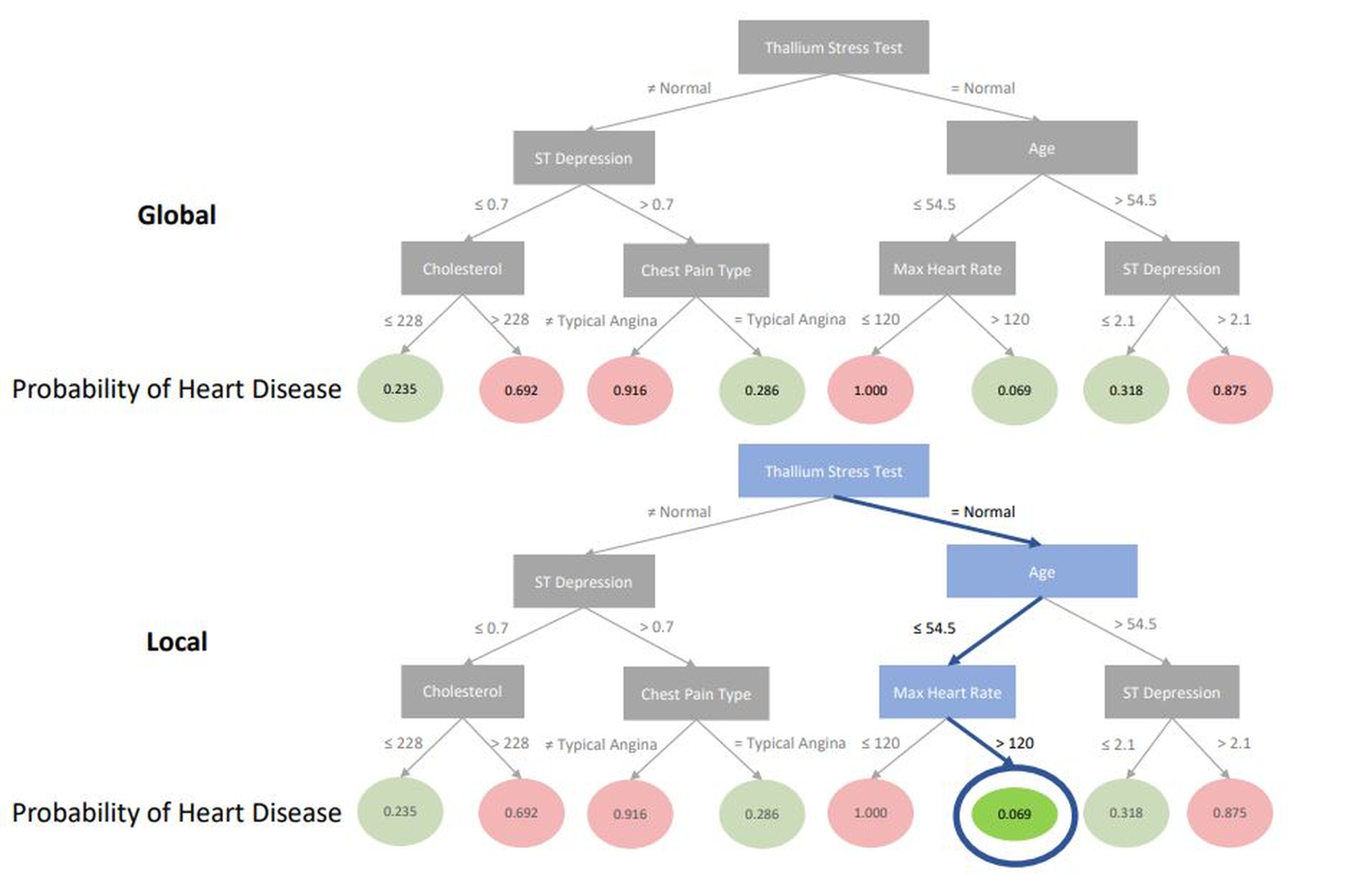}
\caption{A model using random forests predicts heart disease by analyzing both local and global decision trees. The global diagram starts by examining whether a patient's thallium stress test results are normal. If the test shows a problem, the model looks at the patient's ST depression next. The local graphic shows the specific pathway a patient took in the model, explaining the reasons for their individual prediction. For example, a patient under 54.5 years old, with a maximum heart rate that is high and normal thallium stress test results, has a very low chance of having heart disease \cite{petch2022opening}.
}
\label{eai5}
\end{figure}

Figure \ref{eai5} shows a decision tree created from the predictions of a random forest model. This global tree diagram illustrates how the random forest works overall. By following a patient's path through the tree, individual predictions can be examined. This type of explanation is beneficial because it clarifies both the general functioning of the model and the reasoning behind specific predictions. Decision trees are suitable for fields like cardiology because they use rule-based reasoning similar to clinical decision guidelines.

\begin{figure}
\centering
\includegraphics[height= 7 cm]{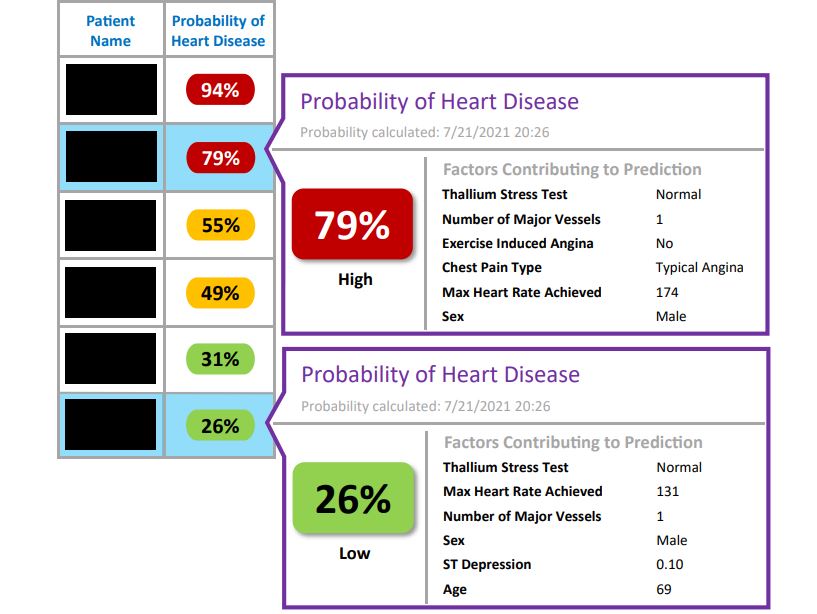}
\caption{Heart disease prediction explanation produced using Local Interpretable Model-Agnostic Explanations (LIME). This illustration shows how clinical decision assistance can be integrated into an Epic electronic health record by means of a local explanation utilizing the LIME algorithm. To help clinicians identify patients who are likely to be at a high risk of heart disease, probabilities are color-coded. To improve the predictability and actionability of the results for doctors, the clinical factors that are most significant to the prediction are shown on the right \cite{petch2022opening}. }
\label{eai6}
\end{figure}

Figure \ref{eai6} shows the LIME explanations for our heart failure model's two local predictions. The authors explain how these predictions serve as a clinical decision support tool in Epic, which is an electronic health record designed for doctors (Epic Systems Corporation, Verona, Wisconsin, USA). This kind of explanation helps to clarify the clinical factors that affect each prediction. Importantly, this type of explanation can be added to an EHR, which may improve the practical use of a complex model by making forecasts and clear explanations easy to integrate into clinical work.

For medical AI to work reliably and be widely used, we need to do a lot of research and reach an agreement on important features like explainability, fairness, privacy, and reliability \cite{kim2023requirements}. We must meet clear requirements and standards in any healthcare setting that uses AI, and we need to update these regularly. Additionally, we should establish laws that clarify who is responsible if something goes wrong with a medical AI whether that’s the designers, researchers, healthcare workers, or patients \cite{rajpurkar2022ai}.

\subsection{Explainability and Interpretability of Autonomous Systems}


Explainability and interpretability are crucial concepts in the context of autonomous systems, referring to the ability to understand the decisions and behaviors of these systems. Explainability involves an autonomous system's capacity to provide clear justifications for its actions and choices \cite{atakishiyev2021explainable}. This clarity is essential for fostering acceptance and confidence in AI systems, especially in critical fields such as banking, healthcare, and autonomous vehicles.

While explainability and interpretability are closely related, interpretability focuses more on understanding the internal mechanisms and processes of the autonomous system \cite{alexandrov2017explainable}. An interpretable system offers users insight into the factors and criteria that influence its decision-making, enabling them to grasp how the system arrived at its conclusions.

The research paper The research article \cite{chamola2023review} focuses on trust and dependability in autonomous systems. Autonomous systems have the potential for system operation, rapid information dissemination, massive data processing, working in hazardous environments, operating with greater resilience and tenacity than humans, and even astronomical examination \cite{xu2019explainable}, \cite{yazdanpanah2021responsibility}. Following years of research and development, today's automated technologies represent the peak of progress in computer recognition, responsive systems, user-friendly interface design, and sensing automation. 

\begin{figure}
\centering
\includegraphics[height= 5.5 cm]{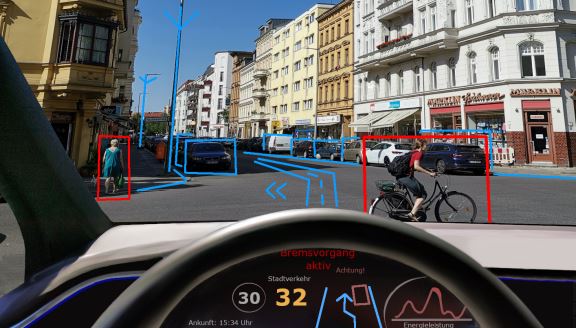}
\caption{An automated vehicle that provides a clear and understandable rationale for its decisions at that moment serves as a prime example of explainable AI in automated driving \cite{chamola2023review}. }
\label{eai7}
\end{figure}

According to \cite{atakishiyev2021explainable}, the global market for automotive intelligent hardware, operations, and innovation is projected to grow significantly, increasing from $1.25 billion in 2017 to $28.5 billion by 2025. Intel's research on the expected benefits of autonomous vehicles indicates that implementing these technologies on public roads could reduce annual commute times by 250 million hours and save over 500,000 lives in the United States between 2035 and 2045 \cite{atakishiyev2021explainable}. Modern cars utilize artificial intelligence for various functions, including intelligent cruise control, automatic driving and parking, and blind-spot detection (Figure \ref{eai7}).

Authors \cite{chamola2023review} describe the challenges of autonomous systems, like, people sometimes tend to be overly excited about the potential of new ideas and ignore, or at least appear to be unaware of, the potential drawbacks of cutting-edge developments. Even in the early stages of robotics and autonomous system implementation, humanity preferred to put up with faulty goods and services, but they have gradually come to understand the importance of trustworthy and dependable autonomous systems. Numerous examples have demonstrated how operators' use of automation is greatly impacted by trustworthiness.

\begin{figure}
\centering
\includegraphics[height= 5.6 cm]{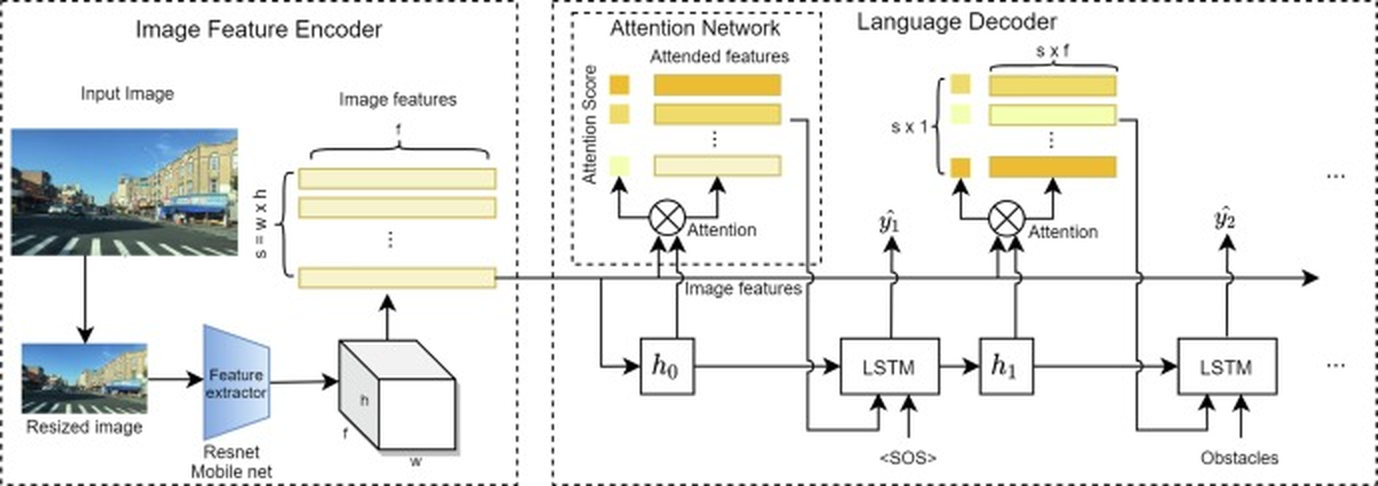}
\caption{The Transformers-based multi-modal deep learning architecture that is being suggested \cite{dong2023did}}
\label{eai8}
\end{figure}

When AI has become prevalent in autonomous vehicle (AV) operations, user trust has been identified as a major issue that is essential to the success of these operations. XAI, which calls for the AI system to give the user explanations for every decision it makes, is a viable approach to fostering user trust for such integrated AI-based driving systems \cite{dong2023did}. This work develops explainable Deep Learning (DL) models to improve trustworthiness in autonomous driving systems, driven by the need to improve user trust and the potential of innovative XAI technology in addressing such requirements. The main concept of this \cite{dong2023did} research is to frame the decision-making process of autonomous vehicles (AVs) as an image-captioning task, generating textual descriptions of driving scenarios to serve as understandable explanations for humans. The proposed multi-modal deep learning architecture, shown in Figure \ref{eai8}, utilizes Transformers to model the relationship between images and language, generating meaningful descriptions and driving actions. Key contributions include improving the AV decision-making process for better explainability, developing a fully Transformer-based model, and outperforming baseline models. This results in enhanced user trust, valuable insights for AV developers, and improved interpretability through attention mechanisms and goal induction.

This research \cite{madhav2022explainable} aims to investigate the integration of XAI into autonomous vehicular systems to improve transparency and human trust. It delves into the functioning of multiple inner vehicle modules, emphasizing the importance of understanding the vehicle's decision-making processes for user credibility and reliability. The main contribution lies in introducing XAI to the domain of autonomous vehicles, showcasing its role in fostering trust, and highlighting advancements through comparative analysis. The output comprises the creation of visual explanatory techniques and an intrusion detection classifier, which show considerable advances over previous work in terms of transparency and safety in autonomous transportation systems.

\subsection{Applications of XAI for Operations in the Industry}

The process industry is a subset of businesses that manufacture items from raw materials (not components) using formulae or recipes.  Given the magnitude and dynamic nature of operations in the process sector, it becomes evident that the next great step ahead will be the capacity for people and AI systems to collaborate to ensure production stability and dependability \cite{hoffmann2021proposal}. AI systems must successfully inform the individuals who share the ecosystem about their objectives, intentions, and findings as the first step toward collaboration. We can hope people to work "with" automation rather than "around" it, thanks in part to the systematic approach to XAI. 

This research \cite{kotriwala2021xai} focuses on XAI applications in the process industry. The research argues that current AI models are not transparent enough for process industry applications and highlights the need for XAI models that can be understood by human experts. The main contribution is outlining the challenges and research needs for XAI in the process industry. The outcome is to develop XAI models that are safe, reliable, and meet the needs of human users in the process industry.

\begin{table}
\caption{Examples of AI applications in process industry operations, including pertinent data, users, and procedures. (RNN = Recurrent Neural Network; KNN = K-Nearest Neighbor; ANN = Artificial Neural Network; SVM = Support Vector Machine; SVR = Support Vector Regression; RF = Random Forest; IF = Isolation Forest) \cite{kotriwala2021xai}}
\label{tab}
\begin{tabular}{|p{1.5cm}|p{2cm}|p{3cm}|p{2cm}|p{1.5cm}|}
\hline
\textbf{Reference} & \textbf{Relevant Data} & \textbf{End Users} & \textbf{Application} & \textbf{AI Methods}  \\ \hline
\cite{mamandipoor2020monitoring}, \cite{cecilio2014nearest}, \cite{banjanovic2017neural} & Process signals  & Operator, Process Engineer, Automation engineer  & Process monitoring & RNN, KNN \\ \hline
\cite{ruiz2001online}, \cite{yelamos2007simultaneous}, \cite{lucke2020fault} & Process signals, Alarms, Vibration  & Process engineer, Automation engineer, Operator, Maintenance engineer  & Fault diagnosis & ANN, SVM, Bayes Classifier  \\ \hline
\cite{dorgo2018learning}, \cite{giuliani2019flaring}, \cite{carter2018application} & Process signals, Acoustic signals  & Operator  & Event prediction & ANN \\ \hline
\cite{desai2006soft}, \cite{shang2014data}, \cite{napier2017isamill} & Process signals  & Operator  & Soft sensors & SVR, ANN, RF \\ \hline
\cite{amihai2018industrial}, \cite{amihai2018modeling}, \cite{kolokas2020fault} & Vibration, Process signals  & Operator, Maintenance engineer, Scheduler  & Predictive maintenance  & RNN, IF  \\ \hline
\end{tabular}
\end{table}

Table \ref{tab} shows examples of AI applied to operational activities in the process industry. This table should give an idea of the breadth of use cases, users, relevant data sources, and applicable AI methodologies; however, it is not intended to be a full or systematic examination.

\section{Future of Trustworthy (XAI)}\label{sec2}

\begin{figure}
\centering
\includegraphics[height= 7 cm]{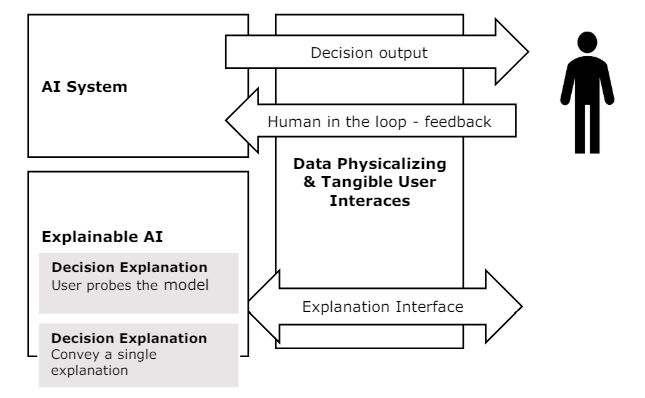}
\caption{Assessing the user's interaction with XAI \cite{thalpage2023unlocking}. }
\label{eai2}
\end{figure}

Figure \ref{eai2} illustrates the precise location of each XAI domain and its relationship with the human user. According to \cite{abdul2018trends}, many explanations of AI systems tend to be static and convey only a single message. However, explanations alone do not facilitate true understanding \cite{adadi2018peeking}. To enhance comprehension, users should have the ability to explore the system through interactive explanations, as most existing XAI libraries currently lack options for user engagement and explanation customization. This represents a promising avenue for advancing the field of XAI \cite{adadi2018peeking} and \cite{abdul2018trends}. Additionally, various efforts have been made to improve human-machine collaboration by moving beyond static explanations.

Explainable AI offers a way to improve how people interact with AI systems. As AI technology grows, it is crucial to ensure that these systems are accountable and transparent. XAI helps by clarifying how AI models work and building trust among users.

We can expect many new developments in XAI. These include making AI models more open, focusing on human-centered designs, ensuring compliance with regulations, and creating hybrid AI systems. XAI will prioritize designs that are easy for users to understand and provide clear explanations. This clarity will help build trust and encourage more people to use AI systems.

Regulatory frameworks are likely to require the use of XAI in important areas to ensure accountability and transparency. Future XAI systems will need to be sensitive to context and provide interactive explanations. This will allow people to engage with AI decisions in real time and adapt to different situations. We must also work to improve digital literacy and tackle ethical issues to ensure that AI systems follow moral principles and society’s values, making XAI technologies accessible to everyone. The success of XAI depends on its ability to bridge the communication gap between AI systems and human users, which encourages cooperation, mutual respect, and trust in an increasingly AI-driven world.

This study \cite{saeed2023explainable} offers a thorough analysis of XAI, focusing on two primary areas of inquiry: general XAI difficulties and research directions, as well as ML life cycle phase-based challenges and research directions. In order to shed light on the significance of formalism, customization of explanations, encouraging reliable AI, interdisciplinary partnerships, interpretability-performance trade-offs, and other topics, the study synthesizes important points from the body of existing literature. The primary contribution is the methodical synthesis and analysis of the body of literature to identify important problems and future directions for XAI research \cite{saeed2023explainable}. The research offers a thorough review of the current state of XAI research and provides insightful information for future studies and breakthroughs in the area by structuring the debate around general issues and ML life cycle phases. The primary finding of the study is the identification and clarification of 39 important points that cover a range of issues and potential avenues for future XAI research. The importance of conveying data quality, utilizing human expertise in model development, applying rule extraction for interpretability, addressing security concerns, investigating XAI for reinforcement learning and safety, and taking into account the implications of privacy rights in explanation are just a few of the many topics covered by these points. Furthermore, the paper indicates directions for further research and application by highlighting the potential contributions that XAI may make to a number of fields, including digital forensics, IoT, and 5G.

\begin{figure}
\centering
\includegraphics[height= 7 cm]{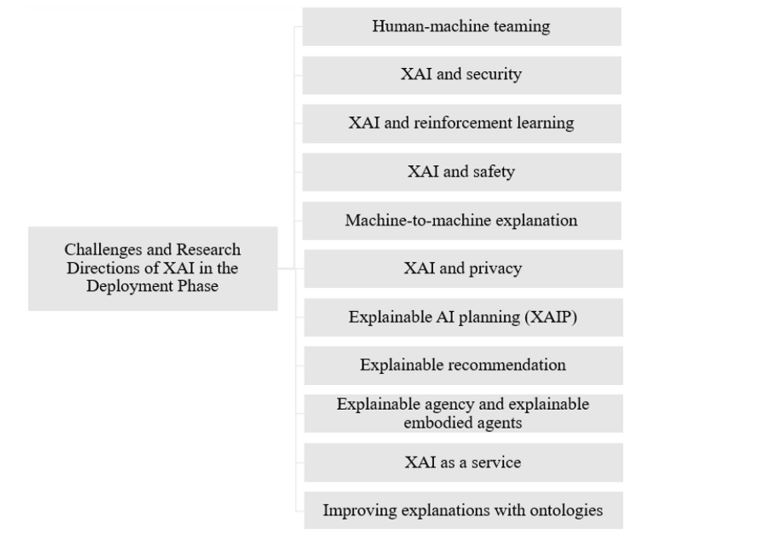}
\caption{Issues and Future Research Paths for XAI throughout its Deployment Stage \cite{saeed2023explainable}. }
\label{eai9}
\end{figure}

Deploying machine learning solutions begins the deployment process and continues until we cease utilizing them, possibly even after that. Figure \ref{eai9} illustrates the XAI research directions and challenges that were explored for this phase.

\section{Conclusions}\label{sec5}

XAI, or Explainable Artificial Intelligence, is becoming important in many industries because it helps solve key challenges with using AI. As AI becomes more common in our daily lives, understanding how it works is essential. XAI provides tools that help people see and understand how AI models make decisions. The main goal of XAI is to make these models easier to understand. It allows people to look inside the "black box" of AI and see what affects its decisions. The paper gives a clear overview of the key parts of XAI. It also discusses three main areas where XAI can be applied. Finally, the authors talk about the challenges of using XAI and suggest possible future directions.

\backmatter

\bmhead{Acknowledgements}

The authors would like to express their sincere gratitude to everyone who encourages and appreciates their scientific work.

\section*{Declarations}

Not applicable



\bibliography{sn-bibliography}

\end{document}